\documentclass[twoside]{article}

\usepackage[accepted]{aistats2022}
\usepackage{longtable}% for long tables
\usepackage{multirow}
\usepackage{bbm}
\usepackage{amsfonts,amsmath}
\usepackage{graphicx}
\usepackage{hyperref}
\usepackage{subfigure}
\usepackage{bm}
\usepackage{bbm}
\usepackage[ruled]{algorithm2e}
\usepackage[round]{natbib}

\begin{document}



\twocolumn[

\aistatstitle{Identifying Incorrect Classifications with Balanced Uncertainty}
\aistatsauthor{Bolian Li \And Zige Zheng \And  Changqing Zhang}
\aistatsaddress{Tianjin University \And  The Chinese University of\\Hong Kong, Shenzhen \And Tianjin University} ]

\begin{abstract}
Uncertainty estimation is critical for cost-sensitive deep-learning applications (i.e. disease diagnosis). It is very challenging partly due to the inaccessibility of uncertainty groundtruth in most datasets. Previous works proposed to estimate the uncertainty from softmax calibration, Monte Carlo sampling, subjective logic and so on. However, these existing methods tend to be over-confident about their predictions with unreasonably low overall uncertainty, which originates from the imbalance between positive (correct classifications) and negative (incorrect classifications) samples. For this issue, we firstly propose the \textbf{distributional imbalance} to model the imbalance in uncertainty estimation as two kinds of distribution biases, and secondly propose \textbf{B}alanced \textbf{T}rue \textbf{C}lass \textbf{P}robability ($BTCP$) framework, which learns an uncertainty estimator with a novel Distributional Focal Loss ($DFL$) objective. Finally, we evaluate the $BTCP$ in terms of failure prediction and out-of-distribution (OOD) detection on multiple datasets. The experimental results show that $BTCP$ outperforms other uncertainty estimation methods especially in identifying incorrect classifications.
\end{abstract}

\section{Introduction}
Deep learning has fundamentally changed the way we co-operate with computing devices. Deep-learning-based applications have emerged into a variety of fields, including computer vision~\citep{forsyth2012computer}, natural language processing~\citep{deng2018deep} and data mining~\citep{han2011data}. However, in some risk-sensitive applications, deploying traditional deep-learning models may bring disastrous outcomes since their predictions are not trustworthy~\citep{floridi2019establishing}, such as disease diagnosis~\citep{ao2020application}, automatic driving~\citep{yasunobu2003auto} and robotics~\citep{davies2000review}. Deep-learning models need to be interpretable~\citep{molnar2020interpretable} and trustworthy~\citep{abdar2020review}. To this end, uncertainty estimation proposes to enable deep-learning models to be self-aware about their predictions. They provide estimated uncertainty for users to choose to trust or not to trust the predictions. The recent uncertainty estimation methods propose to obtain the uncertainty scores from softmax calibration~\citep{hendrycks2016baseline}, Monte Carlo sampling~\citep{gal2016dropout} and subjective logic~\citep{csensoy2018evidential}.

However, there is one limitation for current uncertainty estimation methods. These models tend to be overly confident about all predictions (provide unreasonably low overall uncertainty), even for the incorrect predictions~\citep{mukhoti2020calibrating}. For this issue, we conduct analytical experiments on the origin of this problem, and find that the imbalance problem~\citep{he2009learning} is the primary cause. We also noticed that the traditional imbalance problem as mentioned in \citep{he2009learning} is based on classification task and is modeled with discrete mathematical forms, while in uncertainty estimation, the observed imbalance problem is between correct classifications (with continual low uncertainty groundtruth) and incorrect classifications (with continual high uncertainty groundtruth), which is based on regression task. For this issue, we propose the \textbf{distributional imbalance} to accurately model the imbalance problem in uncertainty estimation, which presents two kinds of distribution biases in the uncertainty distribution. Then, we propose the Balanced True Class Probability ($BTCP$) framework to learn an uncertainty estimator on the Distributional Focal Loss ($DFL$) objective adapted from Focal Loss (FL, \cite{lin2017focal}), which calibrates the uncertainty distribution by end-to-end training. Finally, we evaluate the $BTCP$ in terms of failure prediction and out-of-distribution (OOD) detection on multiple datasets. The experimental results show that $BTCP$ achieves the state-of-the-art performance, and outperforms other uncertainty estimation methods in identifying incorrect classifications.

The main contributions of this paper are summarized as follows:
\begin{itemize}
    \item We propose the distribution imbalance to model the imbalance problem in uncertainty estimation, presenting two kinds of distribution biases in the uncertainty distribution.
    \item We propose the Balanced True Class Probability ($BTCP$) framework to learn an uncertainty estimator in end-to-end training on a novel Distributional Focal Loss ($DFL$) objective.
    \item We evaluate our model in terms of failure prediction and out-of-distribution (OOD) detection on various datasets. Our model achieves the state-of-the-art performance.
\end{itemize}
The rest of the paper is organized as: related works (Sec \ref{sec:related_work}), preliminary (Sec \ref{sec:preliminary}), the introduction of distributional imbalance (Sec \ref{sec:imbalance}), the $BTCP$ framework (Sec \ref{sec:btcp}), experiments (Sec \ref{sec:experiment}), and finally the conclusion and future work (Sec \ref{sec:conclusion}).
\begin{figure*}[ht]
\centering
\includegraphics[width=15cm]{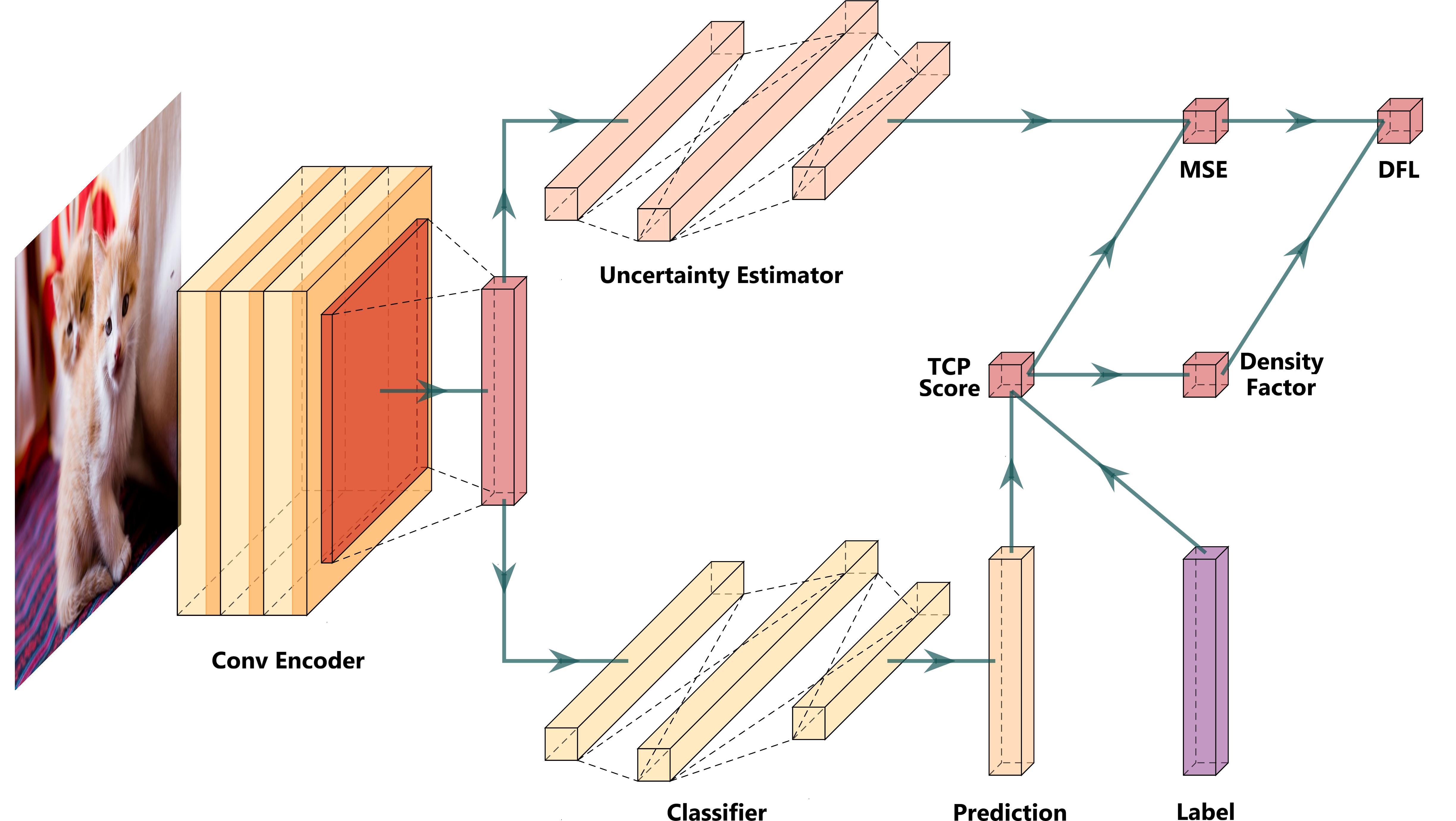}
\caption{Overview of $BTCP$. The classifier and uncertainty estimator share the encoder. In training uncertainty estimator, the output uncertainty is supervised by the $DFL$ objective with groundtruth from the classifier (TCP scores).}
\label{btcp}
\end{figure*}

\section{Related Works\label{sec:related_work}}
In this section, we review the related works, including uncertainty estimation (Sec \ref{subsec:uncertainty_estimation}) and imbalanced learning (Sec \ref{subsec:imbalance_learning}).

\subsection{Uncertainty Estimation\label{subsec:uncertainty_estimation}}
Uncertainty estimation is a critical topic with long history. \cite{blatz2004confidence} summarized traditional methods which are not based on deep learning. \cite{abdar2020review} summarized the techniques, applications and challenges of uncertainty estimation. Recently, MCP~\citep{hendrycks2016baseline} has become a widely used baseline by viewing the maximal value of softmax distribution as confidence. Probabilistic methods with Bayesian modeling obtain the uncertainty by post-process, such as Bayesian neural network~\citep{blundell2015weight}, Monte-Carlo dropout~\citep{gal2016dropout}, prior networks~\citep{malinin2018predictive} and variational inference~\citep{posch2019variational}. Subjective logic~\citep{csensoy2018evidential} employs evidence theory to obtain the uncertainty from Dirichlet distribution. Another approach of uncertainty estimation is based on deep regression, such as TCP~\citep{corbiere2019addressing}. \cite{moon2020confidence} developed a new form of loss function to regularize class probabilities for uncertainty estimation.

\subsection{Imbalanced Learning\label{subsec:imbalance_learning}}
The imbalance problem is summarized in \cite{han2005borderline}. threshold-moving proposed to adjust the decision threshold adaptively. \cite{liu2008exploratory} introduced an under-sampling method with ensemble model. \cite{han2005borderline} developed an over-sampling way via data interpolation. Both of them concentrated on rebuilding balanced training data. \cite{lin2017focal} raised the significance of the scarce classes by applying low weights to the classes with massive samples. \cite{shrivastava2016training} raised the significance of the scarce classes by applying high weights to the classes with few samples.

\section{Preliminary\label{sec:preliminary}}
In this section, we introduce the concepts and definitions of uncertainty in section \ref{subsec:uncertainty} and illustrate the task of failure prediction in section \ref{subsec:failure_prediction}.
\subsection{Uncertainty in Deep Learning\label{subsec:uncertainty}}
Uncertainty in deep learning is the extent to which predictions cannot be trusted. The source of uncertainty is 2-fold: data uncertainty and knowledge uncertainty~\citep{abdar2020review}. There are multiple definitions of uncertainty~\citep{malinin2018predictive,corbiere2019addressing,koh1994identification}. In this paper, we define uncertainty as real number that is complementary with the confidence score, which holds:
\begin{equation}
    u_A+c_A=1,
\end{equation}
where $u_A$ and $c_A$ are the uncertainty and confidence of a prediction $A$ respectively. Moreover, we obtain the uncertainty groundtruth by TCP~\citep{corbiere2019addressing}, which uses the class probability of the labeled class as confidence score:
\begin{equation}
    u_A=TCP(\hat{y}_A,y_A)=1-{\hat{y}_A}^Ty_A,\label{eq:tcp}
\end{equation}
where $\hat{y_A}$ and $y_A$ are the predicted and labeled class probability vectors respectively.

\subsection{Failure Prediction\label{subsec:failure_prediction}}
Failure prediction aims at evaluating the ability for the neural network to recognize when its predictions are wrong~\citep{corbiere2019addressing}. The classification results of a neural network are the groundtruth of failure prediction. The uncertainty is used to predict whether the classification result is correct. When uncertainty is lower than a given threshold: $u<\tau$, it is predicted as correct classification (positive), and when $u\geq\tau$, it is predicted as incorrect classification (negative). The evaluation of failure prediction is based on 4 pairs of conditions (i.e. true positive, false positive, false negative and true negative), which are shown in the confusion matrix in table \ref{tab:fail_prediction}.
\begin{table}[h]\scriptsize%
\centering
\caption{Confusion matrix of failure prediction.}
\label{tab:fail_prediction}
\begin{tabular}{c|c|c}
\hline
uncertainty & correct classification & incorrect classification \\ \hline
$u<\tau$    & true positive (TP)     & false positive (FP)      \\
$u\geq\tau$ & false negative (FN)    & true negative (TN)       \\ \hline
\end{tabular}
\end{table}

\section{Distributional Imbalance\label{sec:imbalance}}
In this section, we introduce the concept of distributional imbalance in section \ref{subsec:definition_imbalance} and provide visual illustration for it in section \ref{subsec:visualization_imbalance}.
\subsection{Concept of Distributional Imbalance\label{subsec:definition_imbalance}}
\begin{figure}[ht]
    \centering
    \includegraphics[width=6.5cm]{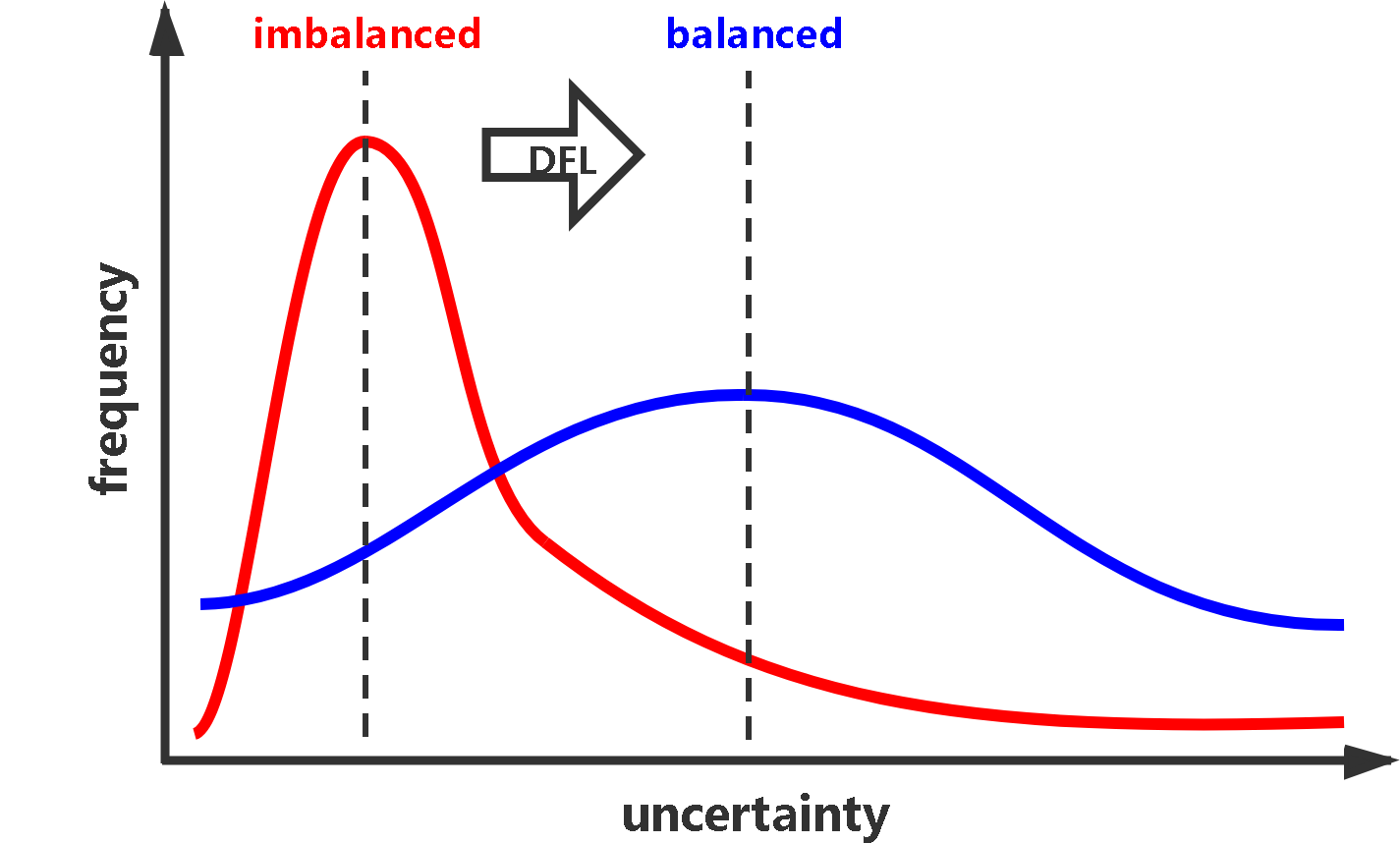}
    \caption{Distributional imbalance.}
    \label{fig:distributional_imbalance}
\end{figure}

The traditional classificatory imbalance~\citep{he2009learning} is based on classification task, in which the different numbers of samples in different classes are the primary cause. We propose the \textbf{distributional imbalance} to accurately model the imbalance problem in uncertainty estimation, considering the continuity of uncertainty scores. Specifically, the imbalanced uncertainty distribution of a neural network would have such 2 kinds of distribution biases:
\begin{itemize}
    \item The means of uncertainty distribution would be low. This is because the uncertainty groundtruth is dominated by massive correct classifications and would be over-fitted on the low uncertainty groundtruth.
    \item The standard deviation of uncertainty distribution would be low. This is due to the unreasonably low overall uncertainty scores, and would lead to the difficulty in distinguishing between easy and hard samples by uncertainty.
\end{itemize}
The above distribution biases are visualized in figure \ref{fig:distributional_imbalance}, where the red distribution suffers from these 2 kinds of biases. Besides, our proposed objective ($DFL$) aims at adjusting the uncertainty distribution from the red one to the blue one. The demonstration of distributional imbalance on real-world datasets is shown in section \ref{subsec:visualization_imbalance}.

The harm of distributional imbalance is that, estimated uncertainty tends to be low even for incorrect classifications. It would cause the neural network to give wrong predictions confidently, which is unacceptable in some risk-sensitive applications like disease diagnosis~\citep{ao2020application}.

\subsection{Visualization of Distributional Imbalance\label{subsec:visualization_imbalance}}
\begin{figure}[ht]
    \centering
    \includegraphics[width=8.2cm]{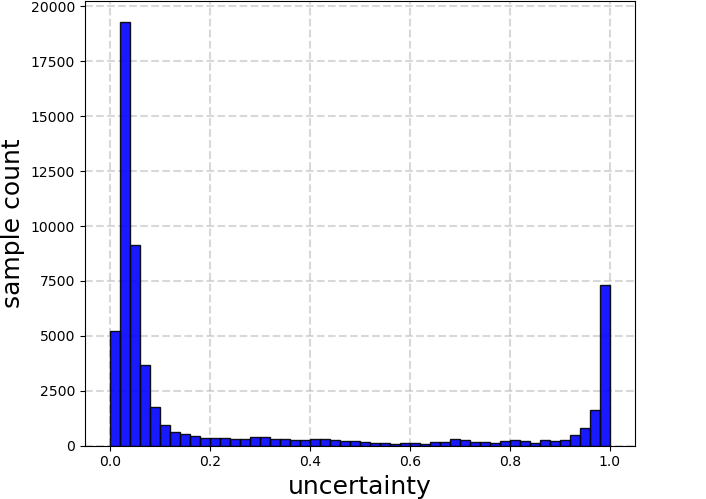}
    \caption{Uncertainty histogram on Fashion-MNIST.}
    \label{fig:hist}
\end{figure}

First, we visualize the TCP uncertainty scores (Eq.\ref{eq:tcp}) on Fashion-MNIST with the uncertainty histogram (50 bins, shown in Fig.\ref{fig:hist}). The uncertainty scores are distributed just as what we talked in section \ref{subsec:definition_imbalance}. The real-world uncertainty distribution suffers from the distributional imbalance with low means and standard deviation.

\begin{figure}[ht]
    \centering
    \subfigure[$BTCP$]{\includegraphics[width=3.9cm]{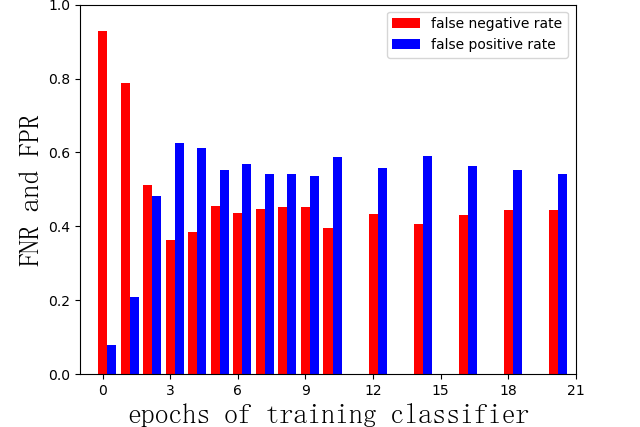}}
    \quad
    \subfigure[TCP]{\includegraphics[width=3.9cm]{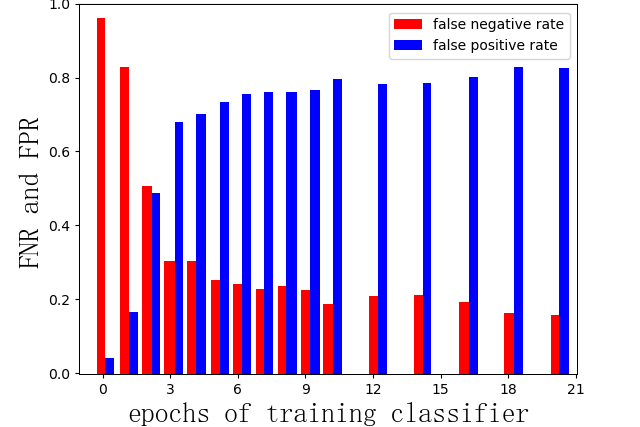}}
    \quad
    \subfigure[OHEM Loss]{\includegraphics[width=3.9cm]{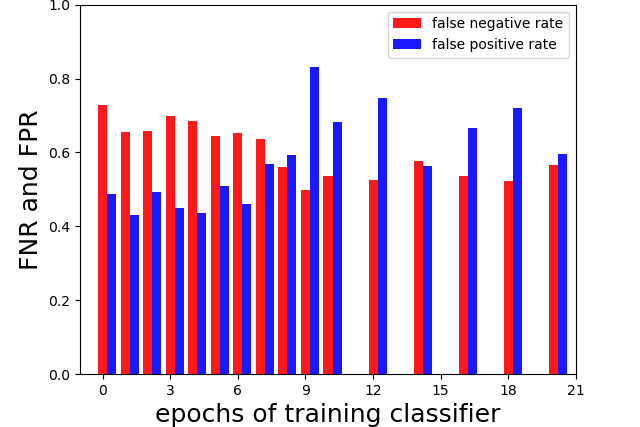}}
    \quad
    \subfigure[thresholding]{\includegraphics[width=3.9cm]{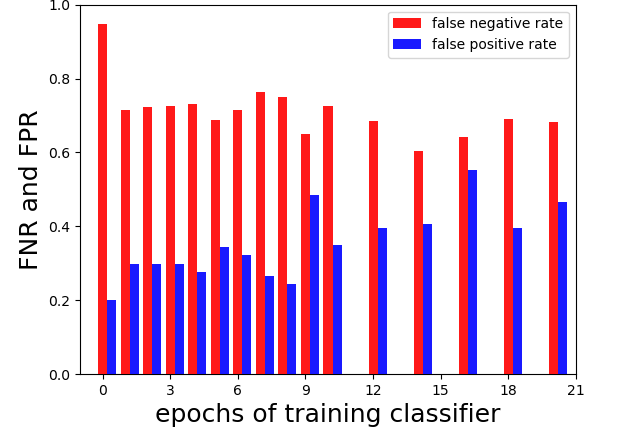}}
    \caption{Visualization of FPR and FNR on Fashion-MNIST.}
    \label{fig:FPRFNR}
\end{figure}

Second, we visualize the false positive rate (FPR) and false negative rate (FNR) of 4 methods in figure \ref{fig:FPRFNR}. The FPR is used to measure how likely incorrect classifications are viewed as correct ones, which is computed by
\begin{equation}
\begin{aligned}
    FPR=1-R_{incorrect}=\frac{FP}{FP+TN},\label{eq:fpr}
\end{aligned}
\end{equation}
and FNR is used to measure how likely correct classifications are viewed as incorrect ones, which is computed by
\begin{equation}
\begin{aligned}
    FNR=1-R_{correct}=\frac{FN}{TP+FN}.
\end{aligned}
\end{equation}
Here, $R_{correct}$ and $R_{incorrect}$ are the recall ratios of correct classification and incorrect classification respectively. We train each model for totally 20 epochs and make a checkpoint at each epoch. The evaluated FPRs and FNRs are based on uncertainty obtained from each checkpoint. Generally, the TCP suffers from high FPR, which is the consequence of distributional imbalance. 

Moreover, the other 3 methods ($BTCP$, thresholding and  OHEM Loss~\citep{shrivastava2016training}) all consider the imbalance problem in uncertainty estimation. Among them, our proposed $BTCP$ outperforms OHEM Loss woth lower FPR, and has better overall performance compared to thresholding with averagely lower FNR and FPR.

\section{Method\label{sec:btcp}}
In this section, we propose a novel Distributional Focal Loss (DFL) in section \ref{subsec:dfl} and introduce the training scheme of uncertainty estimator in section \ref{subsec:learning}. Our discussion about objective and training process are based on the framework in figure \ref{btcp}.

\subsection{Distributional Focal Loss\label{subsec:dfl}}
\textbf{Previous work.} To address the imbalance problem in classification tasks, \cite{lin2017focal} proposed Focal Loss (FL) to adaptively reweight the CrossEntropy objective based on different class volumes. It is defined as
\begin{equation}
    FL(\hat{p}_d)=-(1-\hat{p}_d)^\gamma log(\hat{p}_d),\label{eq:fl}
\end{equation}
where $\hat{p}_d$ is the predicted probability of the $d$-th class, and $\gamma$ is a fixed hyperparameter controlling the sensitivity of reweighting. This objective eliminates the imbalance originating from different class volumes.

\textbf{Motivation.} We are inspired by FL and are motivated to adapt the FL to uncertainty estimation. Given that uncertainty estimation is inherently a kind of regression tasks, and that it suffers from the distributional imbalance (see Sec \ref{sec:imbalance}), we propose a novel Distributional Focal Loss ($DFL$) objective and argue that:
\begin{itemize}
    \item First, we should employ the classic Mean Squared Error (MSE, \cite{wang2009mean}) to compute the unweighted losses of all uncertainty scores, since uncertainty estimation is a kind of regression tasks.
    \item Second, we should adjust the unweighted losses by a density factor similar to Eq.\ref{eq:fl}. We use the density factor to measure the local frequency of given uncertainty scores in the uncertainty distribution.
\end{itemize}

\textbf{Formulation.} Following the above motivations, we define the $DFL$ objective as:
\begin{equation}
    DFL(\hat{u},u)=\left[1-\frac{|N(u)|}{|U|}\right]^\gamma||\hat{u}-u||^2,\label{eq:dfl}
\end{equation}
where $\hat{u}$ is the estimated uncertainty score and $u$ is the groundtruth (TCP score). $N(u)$ is a small continual subset centering at $u$: $u-\epsilon\sim u+\epsilon$, consisting of samples with similar uncertainty. The relative volume of this subset (or density factor): $|N(u)|/|U|$, is a rational approximation of local frequency in the uncertainty distribution. The $DFL$ objective can lead the neural network to focus less on the massive low uncertainty groundtruth, which originates from too many correct classifications.

\subsection{Learning Uncertainty Estimator\label{subsec:learning}}
We train the $BTCP$ framework to learn a balanced uncertainty estimator following the training scheme of \cite{corbiere2019addressing}. First, we train the classifier on CrossEntropy objective with class labels. Second, we train the uncertainty estimator on $DFL$ objective with uncertainty groundtruth (TCP scores). Finally, we fine-tune the entire model by carrying out the above 2 steps simultaneously. The training process of $BTCP$ is summarized in algorithm \ref{al:learning}.
\begin{algorithm}[ht]
    \caption{Training process of $BTCP$.}
    \label{al:learning}
    \#\emph{classifier} \\
    \For {epoch = 1,2,......}{
        Predict the output softmax by classifier; \\ 
        Compute the CrossEntropy loss; \\
        Update the parameters of classifier and encoder;
    }
    \#\emph{uncertainty estimator} \\
    \For {epoch = 1,2,......}{
        Predict the output softmax by classifier; \\ 
        Predict the uncertainty by uncertainty estimator; \\
        Comupte TCP scores as uncertainty groundtruth; \\
        Compute the $DFL$ loss; \\
        Update the parameters of uncertainty estimator;
    }
    \#\emph{fine-tuning} \\
    \For {epoch = 1,2,......}{
        Predict the output softmax by classifier; \\ 
        Predict the uncertainty by uncertainty estimator; \\
        Comupte TCP scores as uncertainty groundtruth; \\
        Compute CrossEntropy and $DFL$ loss, then combine them as a joint loss; \\
        Update the parameters of all modules.
    }
\end{algorithm}

\section{Experiments\label{sec:experiment}}
In this section, we conduct experiments to evaluate the $BTCP$ framework in failure prediction and out-of-distribution (OOD) detection on multiple datasets. We list the specifications of experiments in section \ref{subsec:setup}, and provide detailed quantitative and visual results in section \ref{subsec:quant_eval} and section \ref{subsec:visual_eval} respectively. Our code is available at \url{https://github.com/lblaoke/balanced-TCP}

\subsection{Experimental Setup\label{subsec:setup}}
\textbf{Datasets.} For comprehensive and fair comparison with other uncertainty estiamtion methods, we use 5 real-world image datasets:
\begin{itemize}
    \item \textbf{CIFAR-10} and \textbf{CIFAR-100}\footnote{\url{http://www.cs.toronto.edu/~kriz/cifar.html}} consist of 60000 32x32 colour images. There are 50000 training images and 10000 test images~\citep{krizhevsky2009learning}.
    \item \textbf{SVHN}\footnote{\url{http://ufldl.stanford.edu/housenumbers/}} is a real-world image dataset for developing machine learning and object recognition algorithms, obtained from house numbers in Google Street View images~\citep{netzer2011reading}.
    \item \textbf{MNIST}\footnote{\url{http://yann.lecun.com/exdb/mnist/}} is a database of handwritten digits, and has a training set of 60000 examples, and a test set of 10000 examples~\citep{lecun1998mnist}.
    \item \textbf{Fashion-MNIST}\footnote{\url{https://github.com/zalandoresearch/fashion-mnist}} is a dataset consisting of a training set of 60000 examples and a test set of 10000 examples. Each example is a 28x28 grayscale image, associated with a label from 10 classes~\citep{xiao2017fashion}.
\end{itemize}

\textbf{Metrics.} For failure prediction, we evaluate models on balanced accuracy (BACC, \cite{brodersen2010balanced}), the area under curve (AUC, \cite{mcclish1989analyzing}) and false positive rate (FPR, defined in Eq.\ref{eq:fpr}). Specifically, BACC is computed by the average of recall obtained on each class:
\begin{equation}
\begin{aligned}
    BACC&=(R_{correct}+R_{incorrect})/2 \\
    &=\frac{1}{2}\left[\frac{TP}{TP+FN}+\frac{TN}{FP+TN}\right],
\end{aligned}
\end{equation}
where $R_{correct}$ and $R_{incorrect}$ are the recall ratios of correct classification and incorrect classification respectively.

For OOD detection, we evaluate models on accuracy (ACC) and average uncertainty (AU). The larger AU a model has, the better it is since larger AU shows that this model is less likely to give low uncertainty to OOD data.

\textbf{Compared methods.} We compare the $BTCP$ model with 2 sets of baseline methods: First, uncertainty estimation methods including MCP~\citep{hendrycks2016baseline}, MCDropout~\citep{gal2016dropout}), TrustScore~\citep{jiang2018trust} and TCP~\citep{corbiere2019addressing}; second, imbalanced learning methods such as OHEM Loss~\citep{shrivastava2016training}.

\textbf{Implementation.} The basic framework of $BTCP$ is based on \cite{corbiere2019addressing}, which consists of a classifier, an uncertainty estimator and a shared encder (shown in Fig.\ref{btcp}). To be specific, for all datasets, we build the encoder module on a series of convolutional layers. For classifier and uncertainty estimator, we build them both on fully connected layers. The details of network architecture are listed in the supplementary material. In evaluations, we evaluate each method on each metric for 5 times and report the average performances for fair comparison. We set the decision threshold $\tau$ as $0.5$ by default. Moreover, we recommend to set the hyperparameter $\gamma$ following
\begin{equation}
    \gamma=\frac{1}{12\cdot var(U)},\label{eq:gamma}
\end{equation}
where $var(U)$ is the variance of uncertainty scores. It is rational since we need to make the objective in equation \ref{eq:dfl} sensitive when the variance of uncertainty distribution is low, which indicates strong distributional imbalance. The detailed deduction of equation \ref{eq:gamma} is listed in the supplementary material.

\subsection{Quantitative Evaluations\label{subsec:quant_eval}}
\textbf{Failure prediction.} The quantitative results on failure prediction are shown in table \ref{result}. Our method achieves the state-of-the-art performance on each dataset for almost all metrics. Specifically, for datasets with complex visual details (i.e. SVHN), our method outperforms the others significantly, while for datasets with simple pixel-level details (i.e. Fashion-MNIST), our method outperforms the others slightly. This is due to the lack of complex visual details, which causes the classifier to give less incorrect classifications and therefore has negative effects on methods that do not consider imbalance problem in uncertainty estimation.
\begin{table*}[ht]
\centering
\caption{Overall performance on failure prediction.}
\label{result}
\begin{tabular}{c|c|cccccc}
\hline
Dataset                        & Metric & MCP   & MCDropout & TrustScore & OHEM Loss & TCP            & BTCP           \\ \hline
\multirow{3}{*}{CIFAR-10}      & BACC   & 64.32 & 78.62     & 78.21      & 78.00     & 71.93          & \textbf{78.80} \\
                               & AUC    & 84.23 & 89.08     & 88.47      & 85.71     & 85.60          & \textbf{89.09} \\
                               & FPR↓   & 10.62 & 9.02      & 5.70       & 2.13      & 7.72           & \textbf{2.01}  \\ \hline
\multirow{3}{*}{CIFAR-100}     & BACC   & 72.86 & 75.78     & 74.70      & 77.14     & 77.51          & \textbf{77.80} \\
                               & AUC    & 85.67 & 86.09     & 84.17      & 87.17     & 86.28          & \textbf{88.07} \\
                               & FPR↓   & 7.20  & 4.68      & 4.74       & 3.53      & 3.19           & \textbf{3.11}  \\ \hline
\multirow{3}{*}{SVHN}          & BACC   & 61.26 & 84.55     & 85.35      & 80.27     & 83.86          & \textbf{87.97} \\
                               & AUC    & 93.20 & 92.85     & 92.16      & 90.60     & 93.44          & \textbf{93.66} \\
                               & FPR↓   & 8.66  & 5.60      & 3.74       & 1.16      & 2.41           & \textbf{0.67}  \\ \hline
\multirow{3}{*}{MNIST}         & BACC   & 85.54 & 89.33     & 90.52      & 87.16     & 90.75          & \textbf{91.52} \\
                               & AUC    & 97.13 & 97.15     & 97.52      & 97.29     & \textbf{97.83} & 97.33          \\
                               & FPR↓   & 3.80  & 2.26      & 3.00       & 0.52      & 0.34           & \textbf{0.26}  \\ \hline
\multirow{3}{*}{Fashion-MNIST} & BACC   & 60.63 & 76.65     & 78.16      & 80.78     & 65.16          & \textbf{81.09} \\
                               & AUC    & 82.93 & 87.04     & 87.53      & 87.53     & 87.99          & \textbf{88.36} \\
                               & FPR↓   & 7.87  & 4.87      & 4.64       & 2.78      & 3.93           & \textbf{0.20}  \\ \hline
\end{tabular}
\end{table*}

\textbf{OOD detection.} The quantitative results on OOD detection are shown in table \ref{ood_result}. We use 4 pairs of in-distribution and OOD datasets. Our method ($BTCP$) achieves the state-of-the-art performance on each dataset for almost all metrics. Such results show that the uncertainty of $BTCP$ still performs well on traditional OOD detection tasks while having a balanced distribution.
\begin{table*}[ht]
\centering
\caption{Overall performance on out-of-distribution (OOD) detection.}
\label{ood_result}
\begin{tabular}{c|cc|cc|cc|cc}
\hline
Dataset/OOD & \multicolumn{2}{c|}{CIFAR-10/SVHN} & \multicolumn{2}{c|}{SVHN/CIFAR-10} & \multicolumn{2}{c|}{MNIST/Fashion-MNIST} & \multicolumn{2}{c}{Fashion-MNIST/MNIST} \\ \hline
Metric      & ACC(\%)         & AU↑             & ACC(\%)         & AU↑              & ACC(\%)            & AU↑                 & ACC(\%)            & AU↑                 \\ \hline
MCP         & 43.97           & 0.3652          & 43.56           & 0.3690           & 53.46              & 0.2745              & 38.42              & 0.2146             \\
MCDropout   & 85.86           & 0.5056          & 74.28           & 0.4927           & 92.48              & 0.5147              & 46.09              & 0.3488             \\
TrustScore  & 87.82           & 0.5919          & 79.38           & 0.5166           & 85.64              & 0.4168              & 42.78              & 0.3261             \\
OHEM Loss   & 99.80           & 0.7056          & 85.93           & \textbf{0.6074}  & 93.66              & 0.5826              & 51.97              & 0.4437             \\
TCP         & 99.64           & 0.7190          & 75.18           & 0.5476           & 88.88              & 0.5434              & 47.09              & 0.4146             \\ \hline
BTCP        & \textbf{99.83}  & \textbf{0.7239} & \textbf{86.04}  & 0.5823           & \textbf{99.56}     & \textbf{0.8216}     & \textbf{56.24}     & \textbf{0.4789}    \\ \hline
\end{tabular}
\end{table*}

\subsection{Visual Evalutions\label{subsec:visual_eval}}
\textbf{Uncertainty distribution.} To show the effectiveness of $BTCP$ in addressing distributional imbalance, we visualize the means and standard deviation of uncertainty distribution from 4 methods. We train 4 models on Fashion-MNIST for totally 20 epochs and make a checkpoint at each epoch. Then we obtain uncertainty scores based on these checkpoints. The uncertainty distribution of $BTCP$ has higher means and standard deviation compared with the other 3 methods, which demonstrates that $BTCP$ is the best method in addressing distributional imbalance.
\begin{figure}[ht]
    \centering
    \subfigure[Means]{\includegraphics[width=3.9cm]{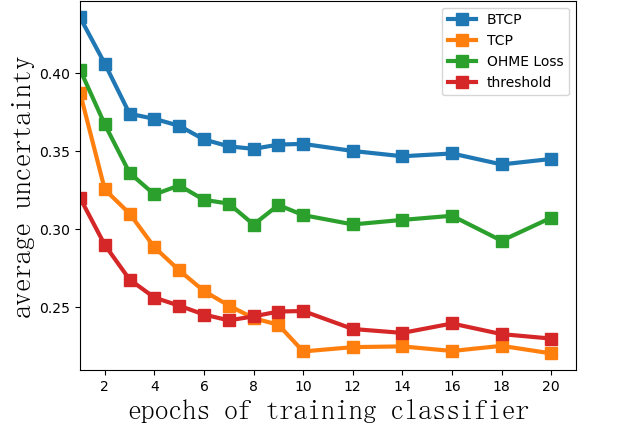}}
    \quad
    \subfigure[Standard deviation]{\includegraphics[width=3.9cm]{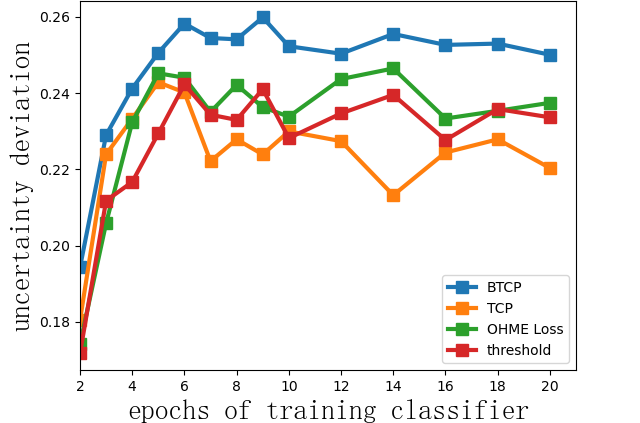}}
    \caption{Visualization of the means and standard deviation of uncertainty distribution on Fashion-MNIST.}
\end{figure}

\textbf{OOD confidence distribution.} We also visualize the confidence distribution of OOD data (red) and in-distribution data (blue) from our method ($BTCP$) and TCP~\citep{corbiere2019addressing}. The in-distribution and OOD datasets are MNIST and fashion-MNIST respectively. Our method outperforms TCP with lower overall confidence (higer overall uncertainty) in the OOD data and distinguishes more clearly between the in-distribution and OOD confidence distributions. Moreover, the in-distribution confidence of our method is also better since it is distributed more scattered and therefore more distinguishable for each sample.
\begin{figure}[ht]
    \centering
    \subfigure[$BTCP$]{\includegraphics[width=3.9cm]{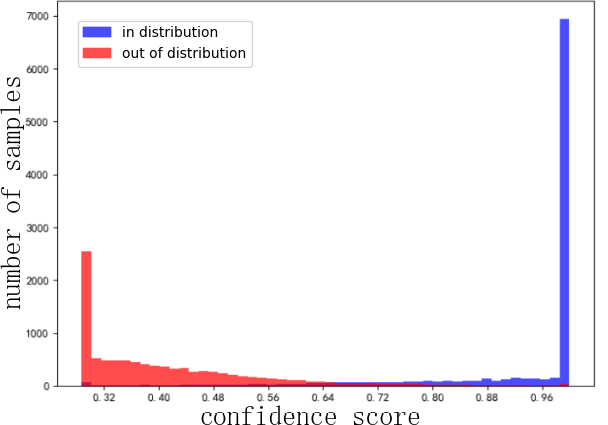}}
    \quad
    \subfigure[TCP]{\includegraphics[width=3.9cm]{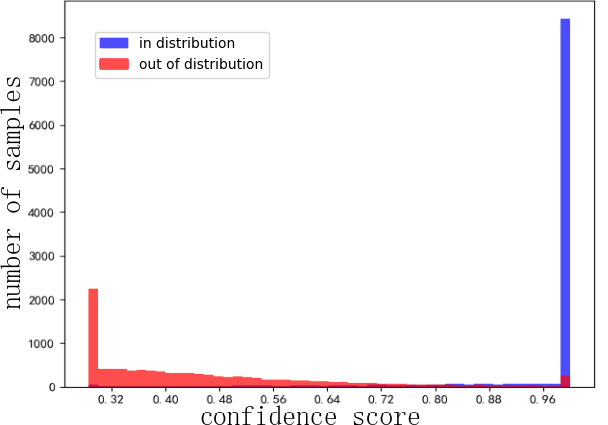}}
    \caption{Visualization of OOD confidence distribution on the MNIST/Fashion-MNIST pair.}
\end{figure}

\section{Conclusion and Future Work\label{sec:conclusion}}
In this paper, we propose the Balanced True Class Probability ($BTCP$) framework. We argue that the distributional imbalance between correct and incorrect classifications consists of two kinds of distribution biases, and is an inevitable problem for uncertainty estimation methods. We model uncertainty estimation as a regression task, propose to learn an uncertainty estimator to obtain the uncertainty scores, and propose to balance the objective for correct and incorrect classifications by a novel Distributional Focal Loss ($DFL$). The experimental results verify the effectiveness of $BTCP$. Our future work will concentrate on introducing probabilistic modeling into the uncertainty estimator and attempting to estimate aleatoric and epistemic uncertainties in a separate way.

\bibliographystyle{plainnat}
\bibliography{main}

\end{document}



% Supplementary material: To improve readability, you must use a single-column format for the supplementary material.
\onecolumn
\aistatstitle{Supplementary Materials}

\section{Implementation}
In designing the structures and hyperparameters of networks, we referred to the original configurations of the TCP project \url{https://github.com/valeoai/ConfidNet}. We use SGD optimizer with learning rate 0.001, fine-tuning learning rate 0.0001, momentum 0.9 and weight decay 0.0001 for each dataset. In testing, we set the batch size of all datasets to 1024. More details of network architecture are listed in table \ref{sm}.
\begin{table}[ht]
\centering
\label{sm}
\caption{Network architectures and hyperparameters.}
\begin{tabular}{c|cccccc}
\hline
Dataset       & \begin{tabular}[c]{@{}c@{}}batch\\ size\end{tabular} & \begin{tabular}[c]{@{}c@{}}classifier\\ epoch\end{tabular} & \begin{tabular}[c]{@{}c@{}}uncertainty\\ estimator\\ epoch\end{tabular} & \begin{tabular}[c]{@{}c@{}}fine-tuning\\ epoch\end{tabular} & \begin{tabular}[c]{@{}c@{}}feature\\ dimension\end{tabular} & encoder \\ \hline
CIFAR-10      & 64                                                   & 15                                                         & 150                                                                     & 15                                                          & 256                                                         & VGG16   \\
CIFAR-100     & 128                                                  & 170                                                        & 170                                                                     & 17                                                          & 512                                                         & VGG16   \\
SVHN          & 64                                                   & 100                                                        & 100                                                                     & 10                                                          & 256                                                         & LeNet   \\
MNIST         & 32                                                   & 50                                                         & 50                                                                      & 5                                                           & 128                                                         & LeNet   \\
Fashion-MNIST & 32                                                   & 20                                                         & 100                                                                     & 20                                                          & 128                                                         & LeNet   \\ \hline
\end{tabular}
\end{table}

\section{Deduction of Hyperparameter Form}
We expect the exponent factor $\gamma$ in $DFL$ vary with the center at $1$. Therefore we need to determine the ``average" value of variance $var(U)$ as a benchmark. Let us consider a situation where all uncertainty scores are distributed uniformly. Assuming that there are $N$ samples in total, and we have:
\begin{equation}
    U=\left\{\frac{i}{N}\right\}_{i=1}^N.
\end{equation}
Then, to calculate the variance of $U$, we need to obtain the means of $U$. When $N$ is large enough, its means is computed by
\begin{equation}
    \overline{U}=\lim_{N\to\infty}\frac{1}{N}\sum_{i=1}^N\frac{i}{N}=\lim_{N\to\infty}\frac{N+1}{2N}=\frac{1}{2},
\end{equation}
which is a constant. Therefore, we can compute its variance, as shown below:
\begin{equation}
    avgvar(U)=\lim_{N\to\infty}\frac{1}{N}\sum_{i=1}^N\left(\frac{i}{N}-\overline{C}\right)^2\approx\int_0^1\left(c-\overline{U}\right)^2dc=\frac{1}{12}.
\end{equation}
Finally, we can satisfy our expectation of $\gamma$ by setting
\begin{equation}
    \gamma=\frac{avgvar(U)}{var(U)}=\frac{1}{12\cdot var(U)},
\end{equation}
which is consistent with Eq.8 in the paper.